\documentclass[runningheads]{llncs}

\usepackage[final]{eccv}

\usepackage{eccvabbrv}

\usepackage{graphicx}
\usepackage{booktabs}
\usepackage{multirow}
\usepackage{bbding}
\usepackage{enumitem}
\usepackage[accsupp]{axessibility}  
\usepackage{pifont} 
\usepackage{pifont}
\newcommand{\cmark}{\ding{51}} 
\newcommand{\xmark}{\ding{55}} 

\usepackage{hyperref}

\usepackage{orcidlink}
\usepackage{mytitlesec}
\titlespacing\section{5pt}{5pt plus 5pt minus 5pt}{5pt plus 5pt minus 5pt}
\titlespacing\subsection{5pt}{5pt plus 5pt minus 5pt}{5pt plus 5pt minus 5pt}
\titlespacing\subsubsection{5pt}{5pt plus 5pt minus 5pt}{5pt plus 5pt minus 5pt}

\begin{document}

\title{DriveVGGT: Calibration-Constrained Visual Geometry Transformers for Multi-Camera Autonomous Driving} 

\titlerunning{Abbreviated paper title}

\author{
Xiaosong Jia\textsuperscript{3*}, 
Yanhao Liu\textsuperscript{1,2*}, 
Yu Hong\textsuperscript{1,2},
Renqiu Xia\textsuperscript{1}, 
Junqi You\textsuperscript{1}, 
Bin Sun, \textsuperscript{4},
Zhihui Hao \textsuperscript{4},
Junchi Yan\textsuperscript{1,\dag}}

\institute{
Shanghai Jiao Tong University \and
Shanghai Innovation Institute  \and
Institute of Trustworthy Embodied AI, Fudan University \and
Li Auto Inc. \\
\textsuperscript{*}Equal Contributions
\textsuperscript{\dag}Corresponding authors
\url{https://github.com/SII-skyboard/DriveVGGT}
}

\authorrunning{F.~Author et al.}


\maketitle

\begin{abstract}

Feed-forward reconstruction has been progressed rapidly, with the Visual Geometry Grounded Transformer (VGGT) being a notable baseline. However, directly applying VGGT to autonomous driving (AD) fails to capture three domain-specific priors: (i) Sparse Spatial Overlap: the overlap among mutli-view cameras is minimal due to $360^{\circ}$ coverage requirements under budget control, which renders global attention among all images inefficient; (ii) Calibrated Geometric Constraints: the absolute distance among cameras is generally accessible for AD data with calibration process before driving. Standard VGGT is unable to directly utilize such information for absolute scale scene reconstruction; (iii) Rigid Extrinsic Constancy: relative poses of multi-view cameras are approximately static, i.e.,  the ego-motion is the same for all cameras.

To bridge these gaps, we propose DriveVGGT, a scale-aware  reconstruction framework that explicitly integrates these priors through three targeted components. First, for the Sparse Spatial Overlap in (i),  we introduce a Temporal Video Attention (TVA) module to process multi-camera videos independently. Second, for Calibrated Geometric Constraints in (ii), a Multi-camera Consistency Attention (MCA) module is designed to directly utilize the calibration information among cameras with a scale head for absolute scale scene reconstruction. Finally, to utilize Rigid Extrinsic Constancy in (iii), we reformulate the decoding process of VGGT into factorized sequential pose head and ego motion head.  On AD datasets, experiments demonstrate that DriveVGGT reduces inference time by 49.3\% while improving depth and pose estimation compared to vanilla VGGT in long-sequence scenarios. It consistently outperforms recent SOTA variants. Meanwhile, extensive ablation studies verify the effectiveness of each devised module.
  \keywords{Feed-forward reconstruction \and Autonomous driving  \and Consistency relationships}
\end{abstract}

\section{Introduction}

\begin{figure}[tb!]
\centering
\includegraphics[width=\linewidth]{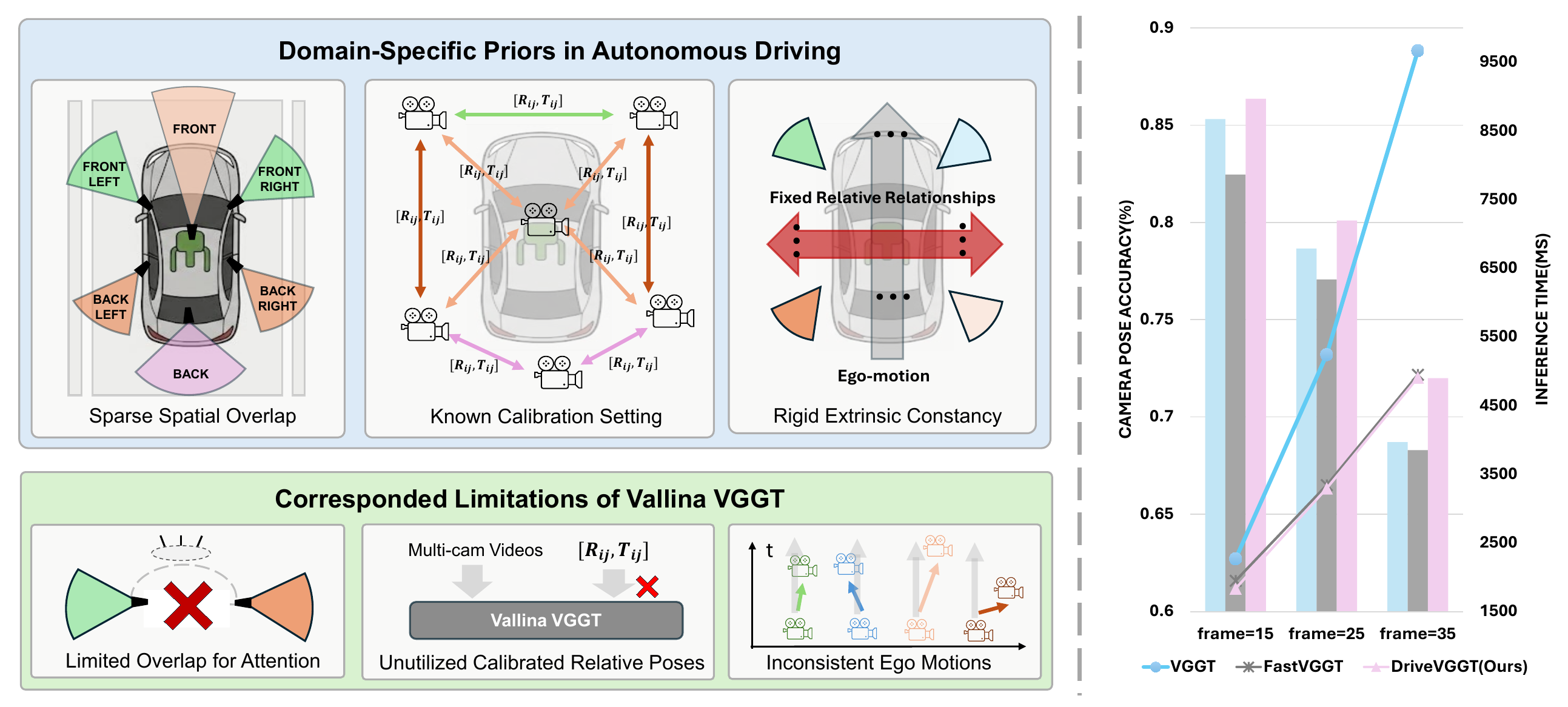}
\caption{
\textbf{Model Comparisons.} By addressing the domain-specific priors in autonomous driving that expose the corresponded limitations of Vallina VGGT, DriveVGGT achieves superior camera pose accuracy and lower inference time. }
\vspace{-5mm}
\end{figure}
\label{sec:intro}


Recently, the field has witnessed a paradigm shift from traditional optimization-based pipelines~\cite{agarwal2011building,frahm2010building,liu2024robust,schonberger2016structure,wu2013towards,lindenberger2021pixel} toward feed-forward, transformer-based architectures~\cite{dust3r,wang20253d,wu2025point3r,wang2025continuous,chen2025ttt3r}, where Visual Geometry Grounded Transformer (VGGT)~\cite{vggt} has emerged as a powerful baseline, demonstrating remarkable generalization across diverse environments. However, when applying to autonomous driving (AD), the general designs of VGGT fail to account for the unique geometric and structural priors inherent in multi-camera vehicle platforms.

Directly deploying vanilla VGGT for AD data reveals three fundamental limitations~\cite{i6,hu2023planning,jia2023hdgt,wu2301policy}. First, to achieve $360^{\circ}$ situational awareness~\cite{i7} under strict cost constraints as a product for consumers, AD camera suites are typically positioned with \textbf{sparse spatial overlap}~\cite{jia2023driveadapter,wu2022trajectory,jia2024amp}, which renders standard global-attention mechanisms in VGGT inefficient~\cite{you2024bench2driver,yang2025drivemoe}. Notably, the number of cameras per frame and the length of videos could be large for AD applicacation, which usually causes degenerated performance due to the qudratic complexity of global attention~\cite{i8}. Secondly, a high-precision calibration process is usually conducted before driving data collection, which provides \textbf{absolute scale distance among cameras}. Standard feed-forward models like VGGT cannot natively ingest these calibrated geometric constraints~\cite{i11}. This leads to an scale ambiguity, where the model struggles to bridge the gap between normalized latent representations and absolute real-world metrical scale~\cite{vggt}. Thirdly, AD systems are characterized by \textbf{rigid extrinsic constancy}. Despite the vehicle's ego-motion, the relative spatial transformation between cameras remains fixed~\cite{driv3r} during the whole driving process. If we naively let VGGT treat multi-view driving videos as independent collections of images, it fails to exploit such temporal-spatial rigidity.

To sufficiently address these bottlenecks and leverage the structural advantages of autonomous driving (AD) systems~\cite{zhu2024flatfusion}, we propose DriveVGGT, a scale-aware reconstruction framework designed to explicitly bake AD-specific priors into the feedforward architecture. Our approach moves beyond treating multi-view sequences as generic image collections, instead framing 4D reconstruction as a geometry-constrained task. There are three core designs to utilize the aforementioned three priors. First, to handle the Sparse Spatial Overlap and mitigate the quadratic complexity of global attention, we introduce a \textbf{Temporal Video Attention (TVA)} module. By processing multi-camera videos independently in the temporal domain, we capture the rich spatiotemporal continuity of each view while avoiding the computational redundancy of attending to non-overlapping cross-view regions. Second, to incorporate Calibrated Geometric Constraints, we design a \textbf{Multi-camera Consistency Attention (MCA)} module. This module integrates known extrinsic calibrated information into the feedforward process and outputs via a dedicated scale head, enabling the model to transition from relative, normalized depth to absolute, metric-scale scene reconstruction. Finally, we exploit Rigid Extrinsic Constancy by \textbf{reformulating the decoding process}. Rather than predicting independent poses for every image, we pool the camera tokens by two factorized dimensions and then feed them into a sequential pose head and an ego-motion head respectively. This factorization ensures that the predicted camera trajectories remain physically consistent with the rigid mounting of the sensor suite. 
 
On the nuScenes dataset~\cite{i10} with  6 low-overlap cameras on the vehicle, DriveVGGT demonstrates superior performance over vanilla VGGT and other recent SOTA baselines. Notably, by decoupling the attention mechanism and factorizing the pose heads, our model achieves a 49.3\% reduction in inference latency, providing a scalable and efficient solution for long-sequence reconstruction for autonomous driving.

In summary, our contributions are as follows:
\begin{itemize}[leftmargin=10pt, topsep=0pt, itemsep=1pt, partopsep=1pt, parsep=1pt,label=$\bullet$]
    \item We propose DriveVGGT, a scale-aware feed-forward reconstruction framework that explicitly integrates the domain-specific priors of autonomous driving.
    \item We introduce a Temporal Video Attention (TVA) module and a Multi-camera Consistency Attention (MCA) module to efficiently handle long-sequence multi-view data while maintaining geometric consistency across cameras.

    \item We reformulate the VGGT decoding process with factorized pose and ego-motion heads, ensuring absolute scale awareness and rigid-body consistency.
    \item  Extensive experiments on AD benchmarks show that DriveVGGT significantly outperforms vanilla VGGT-based methods in both accuracy and speed, reducing latency by nearly half in long-sequence scenarios.
\end{itemize}

\section{Related Works}
\subsection{Scene Reconstruction}
In recent years, feed-forward  reconstruction has been widely researched due to their powerful generalization ability. \cite{dust3r} proposes the first end-to-end 3D reconstruction pipeline to predict images' poses, intrinsics and depths simultaneously. Furthermore, \cite{monst3r} expands 3D reconstruction from static scenes to dynamic scenes; they additionally predict dynamic masks to achieve better performance for dynamic objects. Subsequently, \cite{megasam} implements the learned movement map to recognize motion probability of various objects in seniors exhaustively. To simplify the pipeline of end-to-end reconstruction, VGGT\cite{vggt} proposes a transformer-based feed-forward method to decode various geometric information from image tokens. After that, \cite{stream} achieves streaming reconstruction with VGGT by storing tokens in a memory cache, which can significantly enhance time inference with temporal causal attention. To further optimize VGGT, fastvggt\cite{fastvggt} applies region-based random sampling to improve inference time, which is especially significant when processing with considerable(1000+) image inputs. Furthermore, \cite{faster} implements block-sparse global attention instead of a fully global one to improve model efficiency.

\subsection{Temporal-Spatial Geometry Consistency}

Monocular camera geometry consistency mainly focuses on temporal continuity~\cite{wang2025diffusion,xue2025human,gao2020using,jia2020sentimem}. \cite{wu2025video} utilizes static point clouds to improve the video generation consistency of the diffusion transformer. In the training process, \cite{zhou2025stable} implements an M-in N-out architecture to recover missing images in temporal sequences. Meanwhile, \cite{you2024nvs} proposes a training-free pipeline to consistently generate high-quality novel views. \cite{baisyncammaster} implements a multi-view synchronization module to control the geometry consistency between two different views. \cite{kuang2024collaborative} implements mask attention to maintain consistency of the overlap region in the generation process. In the autonomous driving field, \cite{lu2024seeing} introduces decoupled attention to achieve efficient interaction and eventually keep temporal-spatial geometry consistency among 6-view cameras. Furthermore, to enhance 4D reconstruction stability, recent generative world models~\cite{hu2023gaia, wang2024drivedreamer} incorporate probabilistic state transitions. Additionally, integrating explicit spatial alignments~\cite{chen2024unimlvg} ensures robust multi-view video generation across complex driving scenarios.

\subsection{Position Geometry Representation}
In typical scene reconstruction, estimating 6-D image poses is crucial for achieving better prediction results\cite{hao2025research}. As to the representation of image position, \cite{kong2024eschernet} introduces a novel camera positional encoding method to represent both 4 DoF (object-centric) and 6 DoF camera poses. \cite{xu2024se} implements spherical harmonics as positional encodings, which are specialized equivariant to represent relative poses. \cite{miyatogta} explores a sufficient and effective transformer structure to deal with the specificity of position tokens. \cite{licameras} proposes a novel relative positional encoding technique to process images similar to ROPE\cite{su2024roformer} in attention blocks. When it comes to recent reconstruction works, VGGT\cite{vggt} adds extra image pose tokens to each image and finally decodes as image extrinsics and intrinsics. However, in VGGT, the first frame's camera pose needs to be initialized separately as a position reference for other frames, which may decrease prediction accuracy when feeding images to the model with different orders. Concerned the above problem, \cite{wang2025pi} introduces a fully permutation-equivariant
architecture to eliminate this bias and proposes different orders of images at the same level.


\begin{figure}[tb!]
\centering
\includegraphics[width=\linewidth]{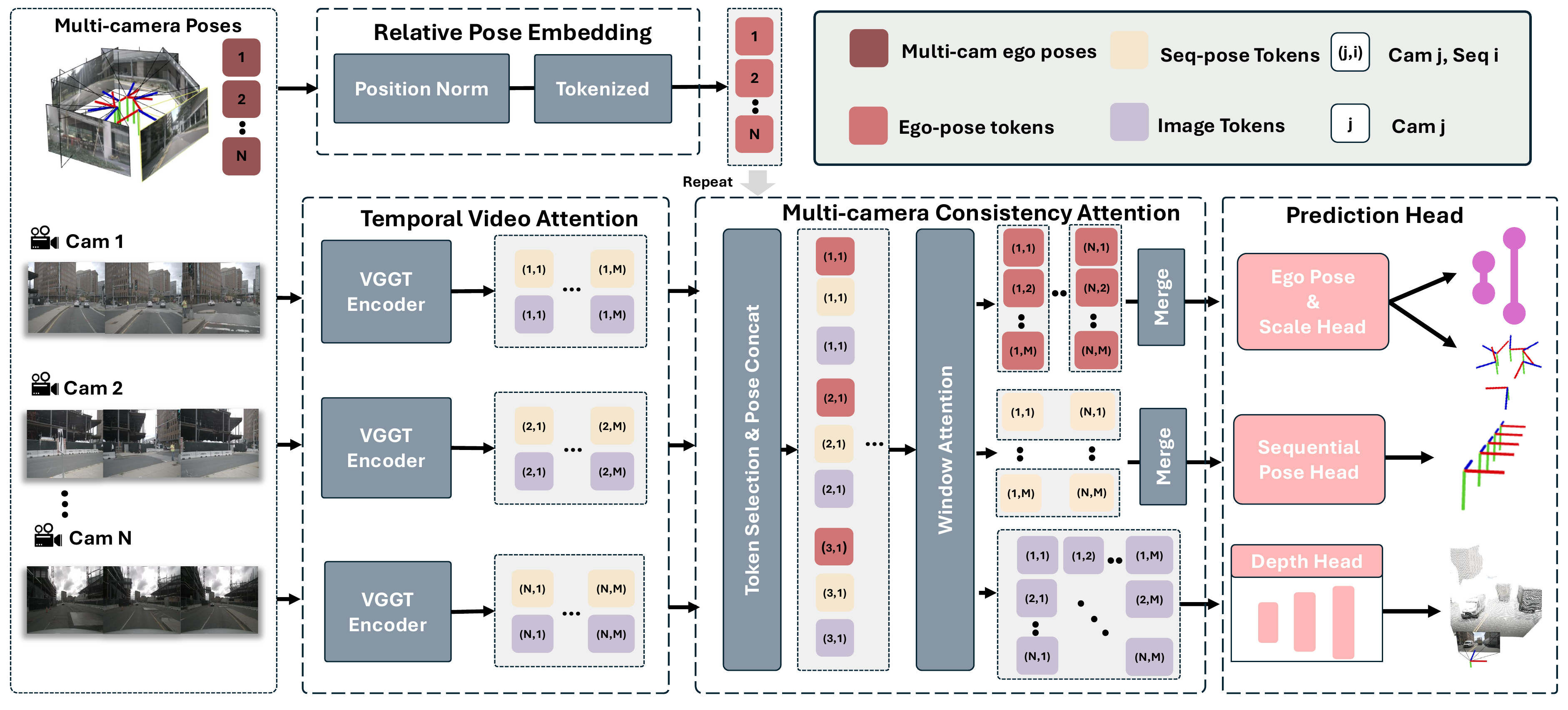}
\caption{\textbf{Overview of DriveVGGT Architecture.} Our framework processes multi-camera AD videos through a three-stage geometric pipeline: \textbf{1) Temporal Video Attention (TVA):} Independent temporal encoders extract spatiotemporal features for each camera stream, mitigating the complexity of global attention. \textbf{2) Multi-camera Consistency Attention (MCA):} Calibrated extrinsic priors are injected into a window-based attention module to resolve scale ambiguity and ensure cross-view geometric consistency. \textbf{3) Factorized Decoding:} The model decouples the geometry into ego-motion and rigid extrinsic heads, enabling high-fidelity, scale-aware reconstruction and consistent camera trajectories across long sequences.}
\label{fig:overal}
\vspace{-4mm}
\end{figure}

\section{Proposed Approach}

The goal of \textbf{DriveVGGT} is to transition from generic feed-forward reconstruction architecture to a scale-aware, geometry-constrained framework tailored for the structural priors of AD. Given a multi-camera AD video sequence consisting of $M$ cameras and $T$ temporal frames, our model aims to predict consistent 3D scene geometry, absolute metric depth, and camera trajectories. As in Fig.~\ref{fig:overal}, the architecture departs from the global-attention paradigm of vanilla VGGT. Instead, it employs a decoupled approach: (i) a Temporal Video Attention (TVA) module to capture intra-camera motion, (ii) a Multi-camera Consistency Attention (MCA) module to fuse cross-view information using calibration priors, and (iii) a Factorized Decoding head that enforces rigid extrinsic constancy.

\subsection{Temporal Video Attention}

In AD scenarios, the high dimensionality of multi-view video data ($M \times T$ images) makes global attention across all frames computationally prohibitive due to its $O((M \times T)^2)$ complexity. Furthermore, the \textbf{Sparse Spatial Overlap} prior suggests that cross-view attention is often redundant.

To this end, we propose Temporal Video Attention (TVA) module to process each camera stream independently. TVA module is proposed to establish the initial geometry relationship among images captured by each camera. These images belong to a streaming video and are suitable for pretrained feed-forward geometry models (like VGGT) to process effective reconstruction results due to the overlap during motion.

Specifically, as to N images, TVA module works as:
\begin{equation} 
  f:\{I_i\}_{i=1}^N \mapsto \{(F_{\mathrm{pos}}(i),F_{\mathrm{depth}}(i))\}_{i=1}^N
\end{equation}
where $I(i)$ is the i-th image with resolution $H\times W$, $f(.)$ is transformer function to process these pictures to tokens. Subsequently, with the help of the decoder head, these tokens can be translated as practical geometry information as:
\begin{equation} 
  \{(Depth(i),Geo(i),...)\}_{i=1}^N = Head(f(\{I_i\}_{i=1}^N))
\end{equation}
where $Depth_i$ is the prediction of the depth map, $Geo(i)$ is image extrinsics, which contain 6 dimensions to indicate the rotation and translation of the image in 3D world, and $F_{pos}(i),F_{depth}(i)$ are geometry tokens of each image.

For multi-camera vehicle platform, unlike global attention in VGGT, TVA only conducts attention for images captured by the same camera. For instance, for M cameras which capture N  images, the function of TVA module is:
\begin{equation} 
  TVA(.) = \{f(.)\}_{i=1}^{M}
\end{equation}
which only aggregate features of each camera, and the outputs is seperate:
\begin{equation} 
  \{(F^{seq}_{pos}(i,j),F_{depth}(i,j))\}_{(i,j)} = TVA(\{I(i,j)\}_{j=1}^{N})
\end{equation}
where $F^{seq}_{pos}(i,j)$ indicates that the camera pose tokens only represent the sequential pose prediction results, which are aligned with the first image of each camera separately. 

\subsection{Relative Pose Embedding}
\label{sec:rel_pose_embed}
A central challenge in feed-forward geometric transformers is the inherent scale ambiguity of the output. To achieve absolute metric-scale scene reconstruction, we need provide the model with physical rulers. For autonomous driving, this could be knwon via the fixed spatial relationships between the $M$ cameras on the vehicle platform obtained via calibration before driving.

Before utilizing the calibration information, we first preprocess the relative extrinsic and intrinsic parameters of each camera $j \in \{1, \dots, M\}$. To ensure numerical stability and align the multi-camera baseline with the latent distribution of the transformer, we perform zero-mean normalization on the camera translation vectors $T_j$:
\begin{equation} 
T_{\text{norm}}(j) = \frac{T_j - \mu_T}{\sigma_T}
\end{equation}
where $\mu_T$ and $\sigma_T$ are the empirical mean and standard deviation of the translations across the sensor rig. Following the geometric encoding paradigm of VGGT \cite{vggt}, we construct a 10-dimensional camera parameter vector $P_{\text{cam}}(j)$ by concatenating the normalized translation, the rotation representation (e.g., quaternions or Euler angles), and the intrinsic parameters:
\begin{equation} 
P_{\text{cam}}(j) = \text{Concat}\left(T_{\text{norm}}(j), R(j), K(j)\right)
\end{equation}
where $K(j)$ denotes the focal length and principal point parameters.

Under the Rigid Extrinsic Constancy prior, these relative poses remain static throughout the driving sequence. Consequently, we only need to compute these embeddings once per sensor configuration. To inject this geometric knowledge into the transformer, we employ a learnable linear projection (or a small MLP) to map $P_{\text{cam}}(j)$ into the high-dimensional latent space of the TVA tokens:
\begin{equation} 
F_{\text{pos}}^{\text{cam}}(j) = \text{Embed}\left(P_{\text{cam}}(j)\right), \quad F_{\text{pos}}^{\text{cam}}(j) \in \mathbb{R}^D
\end{equation}
where $D$ is the embedding dimension of the transformer. These $F_{\text{pos}}^{\text{cam}}(j)$ tokens serve as a static geometric reference, allowing the subsequent Multi-camera Consistency Attention (MCA) module to reason about the relative distance between views in physical units. By explicitly embedding the calibration data, we enable the model to resolve the scale of the scene and output metrically accurate depth maps and camera trajectories.

\subsection{Multi-Camera Consistency Attention}
The features extracted by the TVA module are do not have the multi-veiw information. The captured motion is relative to each camera's specific starting frame and lack a shared metric scale. This results in two primary issues: (i) \textbf{coordinate disconnect}, where each camera trajectory is rooted in an independent local origin, and (ii) \textbf{scale inconsistency}, caused by the lack of cross-view geometric grounding to unify the scale among different cameras.

\begin{figure}[tb!]
\centering
\includegraphics[width=0.85\linewidth]{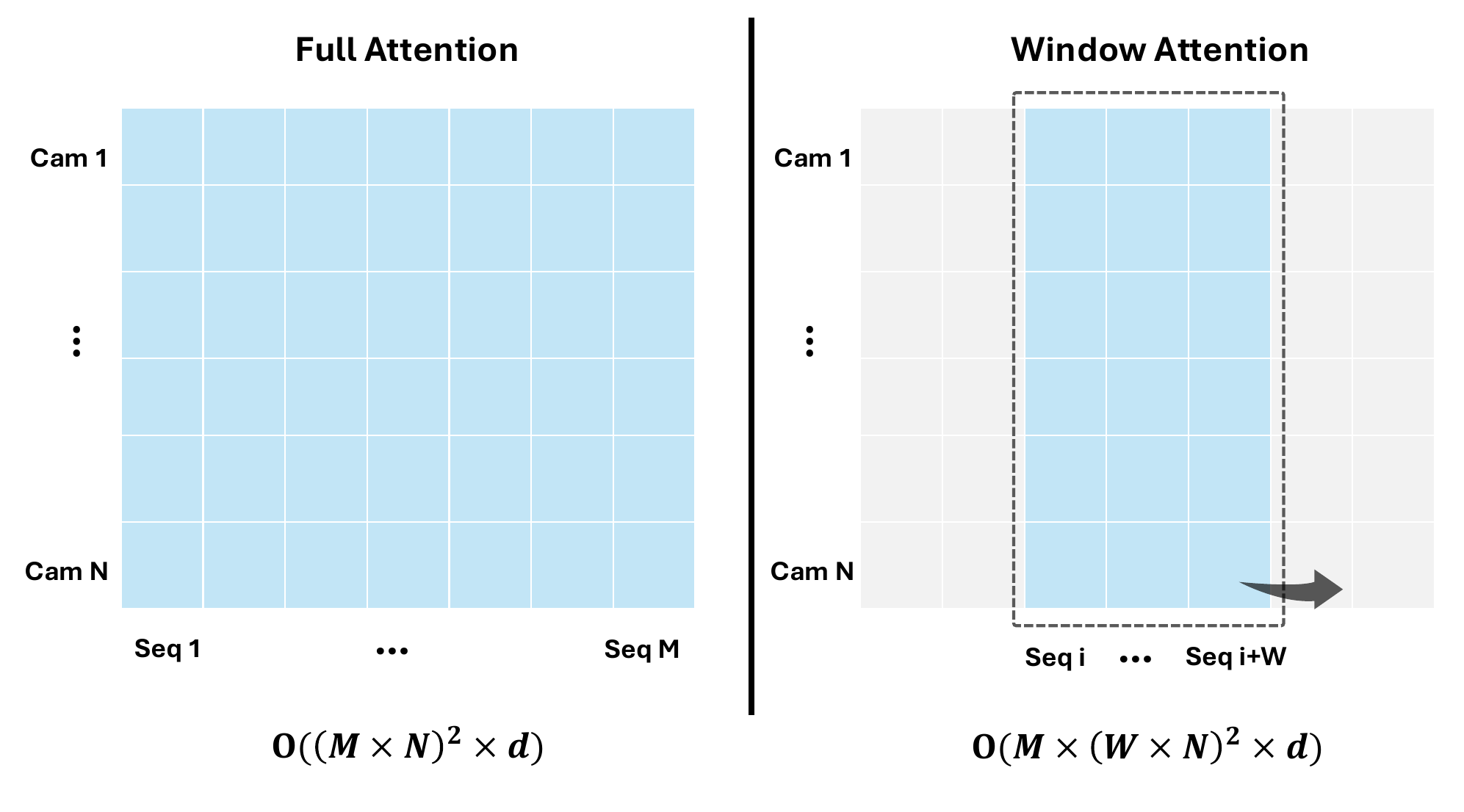}
\caption{\textbf{Window Attention in Multi-Camera Consistency Attention}.}
\label{fig:mca-atten}
\vspace{-4mm}
\end{figure}

To resolve these ambiguities, we propose the \textbf{Multi-camera Consistency Attention} (MCA) module. As in Fig.~\ref{fig:mca-atten},  MCA module utilizes a spatiotemporal windowing strategy to unify these isolated representations into a consistent global 4D ones while maintaining linear complexity relative to sequence length.

\noindent\textbf{Geometric-Visual Token Fusion.}
Before performing cross-view attention, we must ground the visual features in the physical rig geometry. We initialize the MCA tokens by fusing the spatiotemporal features from the TVA module with the calibrated relative pose embeddings described in Sec.~\ref{sec:rel_pose_embed}.

To optimize computational efficiency, we only process tokens from four task-relevant layers of TVA. For a frame $t$ and camera $m$, the input token $\mathcal{F}_{\text{token}}(t,m)$ is constructed via concatenation:
\begin{equation} 
  F_{\text{token}}(t,m) = Concat(F_{pos}^{cam}(m),F^{seq}_{pos}(t,m),F_{depth}(t,m))
\end{equation}
where $\mathcal{F}_{\text{pos}}^{\text{cam}}(m)$ serves as the geometric anchor that informs the model of the camera's fixed position on the vehicle, effectively providing the necessary priors to resolve metric scale.

\noindent\textbf{Spatiotemporal Window Attention.}
Global attention across all $M \times T$ images is computationally exhaustive and often introduces noise from temporally distant frames. Leveraging the Sparse Spatial Overlap prior, we implement a spatial-temporal Window Attention mechanism, which constrains the attention field to a local temporal window of size 2k+1 (covering the preceding, current, and subsequent frames) across all $M$ cameras:
\begin{equation} 
  \{F_{}(i,j)\}_{(i,j)}=\text{Attention}^{i}(\{ \{ F_{token}(i,j)\}_{1}^{M}\}_{i-k}^{i+k})
\end{equation}
By aggregating features from all viewpoints within a local temporal neighborhood, the MCA module can identify cross-camera correspondences and enforce geometric consistency without the quadratic overhead of a full global search. 

\noindent\textbf{Rigidity-Based Feature Aggregation}
To satisfy the Rigid Extrinsic Constancy prior, i.e., the fact that the sensor rig does not deform over time,  we decouple the estimation of movement of the vehicle from the mounting of the cameras through factorized pooling:

\textbf{Ego-Motion Pooling:} To estimate the ego vehicle trajectory, we average the sequential pose tokens across all $M$ cameras for each frame $t$:
\begin{equation} 
F^{seq}_{agg}(t)=\frac1M\sum _{m=1}^MF^{seq}_{}(t,m)
\end{equation}

\textbf{Extrinsic Aggregation:} To obtain a stable estimate of the fixed camera mounting as auxiliary constraint signals, we average the relative pose tokens across the entire temporal sequence of $T$ frames:
\begin{equation} 
    F^{rel}_{agg}(m)=\frac1T\sum _{j=1}^NF^{rel}_{}(i,j)
\end{equation}

This factorization ensures that the final decoding stage produces a physically plausible reconstruction where the cameras move in unison, anchored by the absolute scale provided by the calibrated extrinsics.

\subsection{Prediction Heads}

\textbf{Camera Pose Head:} Following the VGGT architecture, we reuse the camera head in VGGT to decode relative poses and sequential poses separately. Specifically, the Relative Pose Head outputs the time-invariant relative camera poses, which contain camera relative extrinsic and intrinsic. Meanwhile, the Sequential Pose Head proposes the extrinsic parameters of each video camera. Then the two results are aggregated together and propose the camera pose in global axis:
\begin{equation} 
    \mathbf{G}^{global}(i,j)=\mathbf{G}^{seq}(i,j)\times \mathbf{G}^{rel}(i,j)
\end{equation}
where $\mathbf{G}=\begin{pmatrix}R&T\\0&1\end{pmatrix}$, which are decoded from different camera pose head, and $\times$ indicates matrix multiplication. 

\noindent
\textbf{Depth Head:} As VGGT architecture, we use DPT head~\cite{chen2021dpt} to decode the geometry tokens of each image to depths. It outputs prediction results of depths and depth confidence maps. The DPT head implements 4 refinement sub-modules to gradually decode high-resolution dense depths from spatial-compressed geometry tokens.

\noindent
\textbf{Scale Head:} To transform the normalized geometry information to real-world scale, we predict the scale by comparing real-world relative poses and the predicted normalized relative poses as:
\begin{equation} 
    Scale = Avg\left(\sum_j\frac{T^{rel}_{real}(j)}{T^{rel}_{norm}(j)}\right)
\end{equation}
where T indicates relative-pose translation from $\begin{pmatrix}R&T\\0&1\end{pmatrix}$. After multiplying the scale to depths and translation of camera extrinsics, the final world points' prediction can recover to the real-world scale.
\subsection{Loss}
Following the loss design of VGGT, we implement the depth and camera pose loss to supervise DriveVGGT training process. In general, the total loss is calculated as:
\begin{equation} \setlength{\abovedisplayskip}{0pt} \setlength{\belowdisplayskip}{0pt} \setlength{\abovedisplayshortskip}{0pt} \setlength{\belowdisplayshortskip}{0pt}
    L_{total}=\lambda _1L_{depth} + \lambda _2(L_{rel}+L_{seq})
\end{equation}
where $\lambda _1$ is 0.1, $\lambda _2$ is 1.0.

Similar to VGGT, we calculate $L_{depth}$ as 3 parts, which are depth error, depth grad error and uncertainty map error:
\begin{equation} 
L_{\mathrm{depth}}=\sum_{i=1}^{N}\sum_{j=1}^{M}\left\|\hat{\Sigma}_{ij} \left(\hat{D}_{ij}-D_{ij}\right)\right\|+ 
\left\|{\hat\Sigma}_{ij} \left(\nabla \hat{D}_{ij}-\nabla D_{ij}\right)\right\|-\alpha \log {\hat\Sigma}_{ij}
\end{equation}

Compared to VGGT, we decouple the camera pose to two kinds of posed, and $L_{rel}$ is calculated as:
\begin{equation} 
L^{}_{\mathrm{rel}}=\sum_{j=1}^{M}\left\|{\Sigma}_{j} \left(\hat{\textbf{g}}_{j}-\textbf{g}_{j}\right)\right\|_{\epsilon},
\end{equation}
and $L_{seq}$ is calculated as:
\begin{equation} \setlength{\abovedisplayskip}{0pt} \setlength{\belowdisplayskip}{0pt} \setlength{\abovedisplayshortskip}{0pt} \setlength{\belowdisplayshortskip}{0pt}
L^{}_{\mathrm{seq}}=\sum_{i=1}^{T}\left\|{\Sigma}_{i} \left(\hat{\textbf{g}}_{i}-\textbf{g}_{i}\right)\right\|_{\epsilon},
\end{equation}
where $||.||_{\epsilon}$ using the Huber loss.
\begin{figure}[tb!]
\centering
\includegraphics[width=0.9\linewidth]{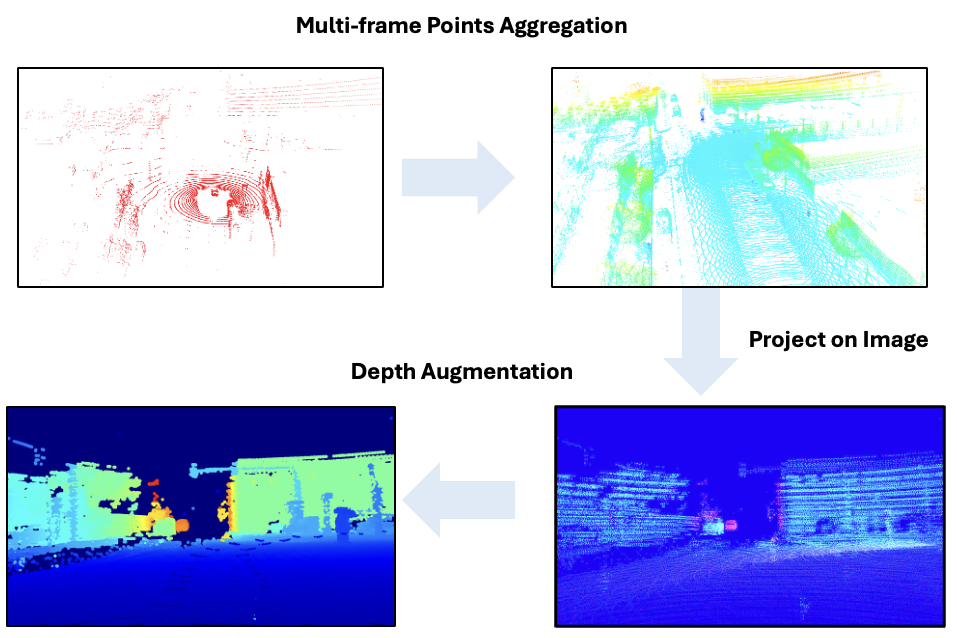}
\caption{\textbf{Two-step Process for Depth Ground-Truth Construction in nuScenes.}}\vspace{-4mm}
\label{fig:nus-depth}
\end{figure}

\section{Experiments}
\subsection{Datasets}
\noindent\textbf{nuScenes Benchmark.} We evaluate DriveVGGT on the large-scale nuScenes dataset~\cite{i10}. The dataset provides $1,000$ driving scenes, each 20 seconds in duration, captured via a sensor suite that includes six cameras providing $360^{\circ}$ coverage, a 32-beam LiDAR, and high-precision localization. Consistent with the Sparse Spatial Overlap prior, the camera setup features minimal shared field-of-view, making it a challenging and representative testbed for our framework. Following standard protocols, we split the data into $700$ scenes for training and $150$ for validation, utilizing keyframes sampled at $2$ Hz.

\noindent\textbf{Dense Depth Ground Truth Generation.} A primary limitation of the nuScenes dataset for feed-forward reconstruction is the relative sparsity of the raw LiDAR returns, which are insufficient for supervised training of high-resolution depth heads. To address this, we implement a two-stage LiDAR Densification Pipeline to generate high-fidelity depth ground truth, as illustrated in Fig.~\ref{fig:nus-depth}:

\begin{enumerate}[leftmargin=10pt, topsep=0pt, itemsep=1pt, partopsep=1pt, parsep=1pt]
\item \textbf{Spatiotemporal Point Aggregation:} We accumulate multiple LiDAR sweeps over a local temporal window to construct a dense global point cloud. For static scene elements, points are registered using the vehicle’s ego-motion. For dynamic actors, we leverage the provided 3D bounding box annotations to aggregate points within the object's local coordinate frame at each timestep, preventing motion blur or "ghosting" artifacts in the resulting point cloud.
\item \textbf{Depth Completion and Augmentation:} The aggregated points are projected onto  image planes using calibrated camera intrinsics. To fill remaining gaps and provide a continuous supervision signal, we apply a depth augmentation algorithm (e.g., bilateral filtering or morphology-based completion).
\end{enumerate}

While this densification process introduces minor noise due to sensor synchronization and projection errors, it yields a significantly richer supervision signal than raw LiDAR. This allows DriveVGGT to learn the complex mapping between visual features and the Calibrated Geometric Constraints required for absolute metric-scale reconstruction.

\subsection{Implementation Details}

For models' inputs, we decrease the initial image resolution of nuScenes from 1600x900 to 518x280, and implement the same changes to images' intrinsics during ground truth generation. Then, like VGGT, we propose scale normalization to the depth map, camera poses to keep the scale consistent, while we additionally use the scale for training. As to the training process, initially, we randomly input 3-10 frame multi-camera images (18-60 images) from scenes for 20 epochs. Each epoch trains 1000 times with 2e-4 learning rate. Then we freeze the aggregator and fine-tune for another 5 epochs with 1e-5 learning rate. For fair comparisons, we train other models under the same way.

\subsection{Pose Estimation}
To compare the pose estimation of the proposed methods with other VGGT-based methods, we test VGGT, StreamVGGT, and fastVGGT on the proposed nuScenes datasets. To illustrate the models' performance on different numbers of images, we set three kinds of image inputs: 15 frames (90 images), 25 frames (150 images), and 35 frames (210 images). Meanwhile, we incorporate relative pose embedding into VGGT and fastVGGT to demonstrate the role of relative poses in these models. Regarding our method, we implement two base geometry transformers to achieve temporal video attention in the TVA module, namely DriveVGGT (VGGT) and DriveVGGT (fastVGGT). The results are shown in Table~\ref{tab:mutli-cam-pose}. Initially, DriveVGGT (VGGT) achieves better performance than other methods, especially in scenes with 210 images. Meanwhile, as to the implementation of camera pose embedding, VGGT and fastVGGT exist performance degradation. However, as to DriveVGGT, the aggregation will improve the accuracy of camera pose estimation, which demonstrates the sufficient use of relative poses.

\begin{table*}[tb!]
\linespread{1.5}
\centering
    \caption{
        \textbf{Multi-camera Pose Estimation on nuScenes dataset.} }
        \vspace{-3mm}
\resizebox{\textwidth}{!}{
\begin{tabular}{cccccccc}
        \toprule
        {\multirow{2}{*}{\textbf{Method}}}&
        {\multirow{2}{*}{\textbf{Relative poses}}}&
        \multicolumn{2}{c}{{\textbf{frame=15}}} &
        \multicolumn{2}{c}{{\textbf{frame=25}}} &
        \multicolumn{2}{c}{{\textbf{frame=35}}} 
        \\
        \cmidrule(r){3-4} \cmidrule(r){5-6} \cmidrule(r){7-8}
        & &
        AUC(30) $\uparrow$ & AUC(15) $\uparrow$ &
        AUC(30) $\uparrow$ & AUC(15) $\uparrow$ &
        AUC(30) $\uparrow$ & AUC(15) $\uparrow$ \\
        \midrule
        VGGT~\cite{vggt}   & \XSolidBrush&
        0.8531 &  \underline{0.7689} &
        \underline{0.7866} & \underline{0.6719}&
        0.6871 & 0.5477 \\
        StreamVGGT~\cite{stream}   & \XSolidBrush&
        0.7005&0.5884&OOM&OOM&OOM&OOM
        
        \\
        fastVGGT~\cite{fastvggt}   &\XSolidBrush &0.8246 & 0.7191 &
        0.7707 & 0.6435 &
        0.6830 & 0.5357 
        \\
        \midrule
        VGGT~\cite{vggt} & \checkmark &  0.8164 &0.7195 &0.7403 & 0.6136 &0.6445 & 0.5002 \\ 
        fastVGGT~\cite{fastvggt}   & \checkmark&
        0.7915&0.6764&
        0.7321&0.5954&
        0.6477&0.4976
        \\
        \textbf{DriveVGGT(VGGT)}  & \checkmark &
         \textbf{0.8635} &  \textbf{0.7706} &
          \textbf{0.8010} & \textbf{0.6778} &
          \textbf{0.7200} &  \textbf{0.5811} \\

        \textbf{DriveVGGT(fastVGGT)}  & \checkmark &
        \underline{0.8534} & 0.7498 &
        0.7844 & 0.6514 &
        \underline{0.6995} & \underline{0.5510} \\
        
        \bottomrule
    \end{tabular}
    }
    \vspace{-5mm}
    \label{tab:mutli-cam-pose}
\end{table*}
\subsection{Depth Estimation}
The comparison of depth estimation is shown in Table~\ref{tab:videodepth}. As the evaluation of camera pose estimation, we test VGGT, StreamVGGT, fastVGGT and DriveVGGT on the nuScenes dataset. As to metric Abs Rel, DriveVGGT(fastVGGT) achieves the best depth estimation performance in scene-35 scenes, which indicates its ability to process long sequence multi-camera videos. StreamVGGT outperforms other methods in frame-15 scenes. 
\begin{table*}[tb!]
    \centering
    \caption{\textbf{Multi-camera Depth Estimation on nuScenes dataset.}}
    \vspace{-2mm}
    \linespread{1.0}
    \resizebox{\textwidth}{!}{
    \begin{tabular}{cccccccc}
        \toprule
        {\multirow{2}{*}{\textbf{Method}}} &
        {\multirow{2}{*}{\textbf{Relative poses}}} &
        \multicolumn{2}{c}{\textbf{frame=15}} &
        \multicolumn{2}{c}{\textbf{frame=25}} &
        \multicolumn{2}{c}{\textbf{frame=35}} \\
        \cmidrule{3-4} \cmidrule{5-6} \cmidrule{7-8}
        & &
        Abs rel $\downarrow$ & $\delta^{3} \uparrow$ &
        Abs rel $\downarrow$ & $\delta^{3} \uparrow$ &
        Abs rel $\downarrow$ & $\delta^{3} \uparrow$ \\
        \midrule
        VGGT~\cite{vggt} & \XSolidBrush &
        0.3666 & 0.8791 &
        \underline{0.3654} & 0.8817 &
        0.3605 & 0.8858 \\
        StreamVGGT~\cite{stream} & \XSolidBrush &
        \textbf{0.3636} & \underline{0.8811} & OOM & OOM & OOM & OOM \\
        fastVGGT~\cite{fastvggt} & \XSolidBrush &
        0.3684 & 0.8782 &
        0.3693 & 0.8794 &
        0.3660 & 0.8825 \\
        \midrule
        VGGT~\cite{vggt} & \checkmark &
        0.3718 & 0.8779 &
        0.3700 & 0.8805 &
        0.3647 & 0.8844 \\
        fastVGGT~\cite{fastvggt} & \checkmark &
        \underline{0.3655} & 0.8784 &
        0.3691 & 0.8795 &
        0.3658 & 0.8826 \\
        \textbf{DriveVGGT(VGGT)} & \checkmark &
        0.3805 & 0.8747 &
        0.3705 & \underline{0.8825} &
        \underline{0.3601} & \underline{0.8892} \\
        \textbf{DriveVGGT(fastVGGT)} & \checkmark &
        \underline{0.3655} & \textbf{0.8854} &
        \textbf{0.3601} & \textbf{0.8894} &
        \textbf{0.3539} & \textbf{0.8935} \\
        \bottomrule
    \end{tabular}}
    \vspace{-1em}
    \label{tab:videodepth}
\end{table*}

\subsection{Inference Time Estimation}
The comparison of inference time is shown in Table~\ref{tab:inference-time}. In general, the proposed method achieves faster inference speed in contrast to VGGT and fastVGGT. The inference time of DriveVGGT(VGGT) is only 50\% of VGGT's in frame-35 scenes. Meanwhile, DriveVGGT(fastVGGT)'s speed is lower than DriveVGGT(VGGT), which is due to the extra token aggregation algorithm in fastVGGT, resulting in a delay in inference time when processing fewer images. 

\begin{table}[tb!]
    \centering
    \caption{\textbf{Multi-camera inference time on nuScenes dataset.}}
    \vspace{-4mm}
    \linespread{1.0}
    \resizebox{0.6\textwidth}{!}{
    \begin{tabular}{cccc}
        \toprule
        {\multirow{2}{*}{\textbf{Method}}} & 
        \multicolumn{3}{c}{\textbf{Inference time (ms) $\downarrow$}} \\
        \cmidrule{2-4}
        & {frames=15} & {frames=25} & {frames=35} \\
        \midrule
        VGGT~\cite{vggt} & 2268 & 5241 & 9666 \\
        StreamVGGT~\cite{stream} & 6916 & OOM & OOM \\
        fastVGGT~\cite{fastvggt} & \underline{1950} & \underline{3341} & \underline{4949} \\
        \midrule
        \textbf{DriveVGGT(VGGT)} & \textbf{1836} & \textbf{3294} & \textbf{4907} \\
        \textbf{DriveVGGT(fastVGGT)} & 2390 & 3823 & 5043 \\
        \bottomrule
    \end{tabular}}
    \vspace{-4mm}
    \label{tab:inference-time}
\end{table}

\subsection{Visualization}
\begin{figure*}[tb!]
\centering
\includegraphics[width=0.82 \linewidth]{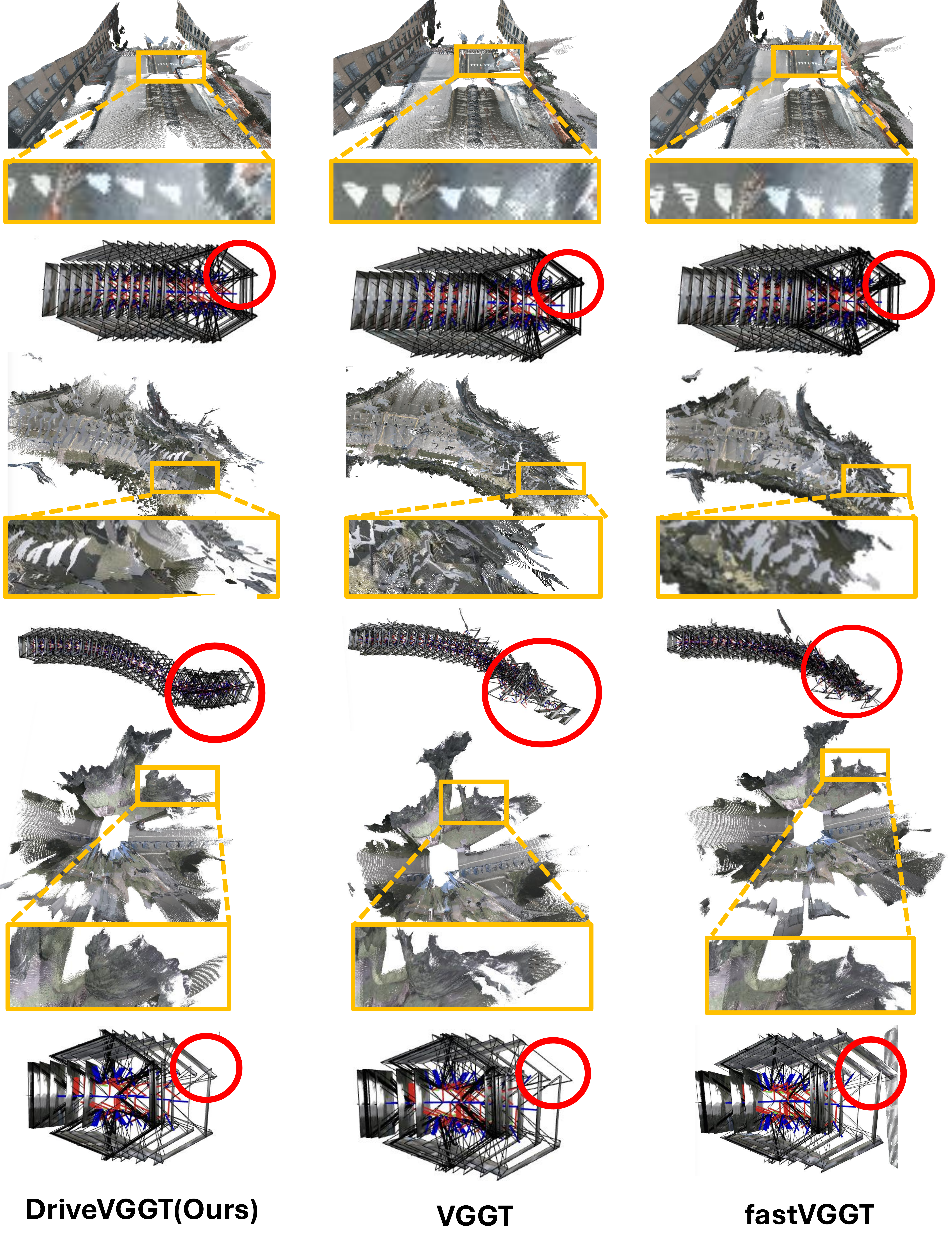}
\caption{\textbf{Qualitative results} of DriveVGGT, VGGT, fastVGGT. We visualize predictions' global points and image poses to compare models' performance comprehensively.}
\vspace{-1mm}
\label{fig:visualization}
\end{figure*}

To qualitatively evaluate the performance of DriveVGGT, we compare its 3D reconstruction and trajectory accuracy against the VGGT and fastVGGT baselines across three representative driving scenarios in Fig.~\ref{fig:visualization}. By back-projecting the predicted metric depth maps into global point clouds using estimated extrinsics for 30-frame sequences (180 images), we observe that while all models handle basic forward motion, fastVGGT frequently exhibits minor rotational misalignments. Crucially, in long-sequence scenarios, both VGGT and fastVGGT suffer from significant temporal drift and accumulated pose errors for frames distant from the temporal origin, resulting in ghosting artifacts and blurred structural outputs in the point clouds. In contrast, DriveVGGT maintains high trajectory stability throughout the sequence, utilizing the TVA and MCA modules to suppress error accumulation and resolve scale ambiguity. This geometric grounding allows our model to produce significantly sharper reconstructions of surrounding environmental details, such as vegetation and roadside infrastructure, confirming the effectiveness of our specialized architectural priors for autonomous driving.

\begin{table}[!t]
    \centering
    \begin{minipage}[t]{0.48\textwidth}
        \centering
        \caption{\textbf{Ablations of Components.}}
        \vspace{-3mm}\label{tab:components}
        \linespread{1.2}
        \renewcommand\arraystretch{1.2} 
        \resizebox{\textwidth}{!}{
        \begin{tabular}{cccc}
            \toprule
            {\multirow{3}{*}{\textbf{Method}}} & \multicolumn{3}{c}{\textbf{frame=25}} \\
            \cmidrule{2-4}
            & AUC(30) $\uparrow$ & Abs rel $\downarrow$ & Time $\downarrow$ (ms) \\
            \midrule
            baseline(TVA) & 0.039 & 0.3711 & \textbf{2052} \\
            +rel pose embed & 0.7855 & 0.3707 & 2098 \\
            \midrule
            DriveVGGT & \textbf{0.8010} & \textbf{0.3705} & 3294 \\
            \bottomrule
        \end{tabular}
        }
    \end{minipage}
    \hfill
    \begin{minipage}[t]{0.48\textwidth}
        \centering
        \caption{\textbf{Ablation of Window Size.}}\vspace{-3mm}\label{tab:windowsizes}
        \linespread{1.2}
        \renewcommand\arraystretch{1.2}
        \resizebox{\textwidth}{!}{
        \begin{tabular}{cccc}
            \toprule
            {\multirow{3}{*}{\textbf{Window size}}}&
            \multicolumn{3}{c}{\multirow{1}{*}{\textbf{frame=25}}}
            \\
            \cmidrule(r){2-4}
            & 
            AUC(30)$\uparrow$ &
            Abs rel$\downarrow$ &
            Time$\downarrow(ms)$ 
            \\
            \midrule
            size=3(Ours) &0.8010 &\textbf{0.3705}& \textbf{3294}\\
            size=5 & 0.8033& 0.3741& 4924\\
            size=7 & \textbf{0.8087}& 0.3744& 7263\\
            \bottomrule
        \end{tabular}
        }
    \end{minipage}
    \vspace{-2mm}
\end{table}

\subsection{Ablation Experiment}
\begin{figure}[tb!]
\centering
\includegraphics[width=0.88\linewidth]{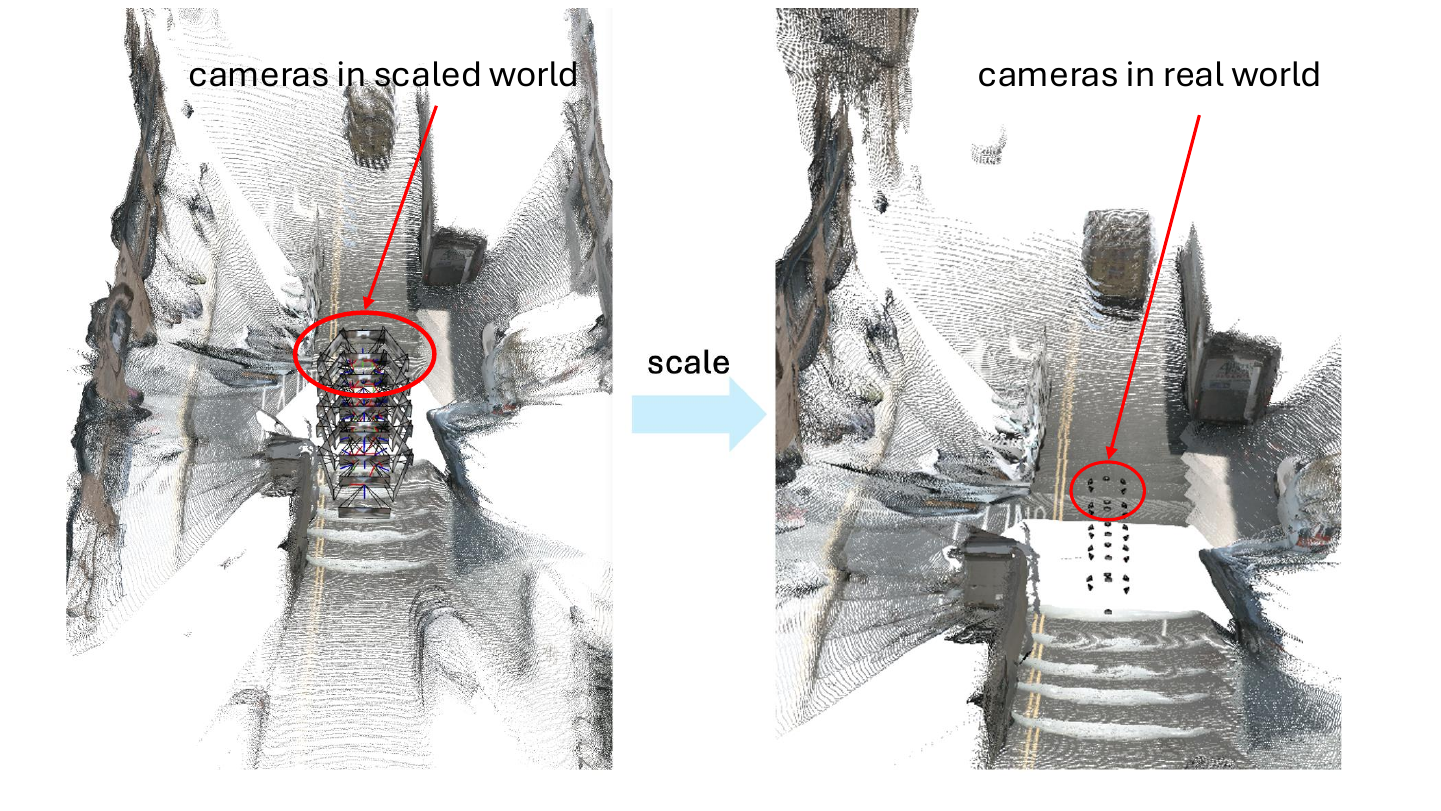}
\caption{\textbf{Comparison between Scaled Scenes and Real Scenes}.}
\label{fig:lst-scale}
\vspace{-5mm}
\end{figure}

To validate the effectiveness of the proposed components, we conduct an ablation study by removing the proposed modules from DriveVGGT, and the detailed evaluation is presented in Table~\ref{tab:components}. The baseline module only utilizes the TVA module to implement attention among images in the video. The test results indicate that the baseline could not handle a multi-camera system as the lack of relative pose representation. After adding relative pose embedding, the model can output a correct pose prediction of the multi-camera system. We evaluate the influence of window size for window attention in Table~\ref{tab:windowsizes}. Where we find that $k=3$ is enough, validating the unnecessity of conducting global attention. To evaluate the effectiveness of the scale head, we compare the depth prediction results with the ground truth using two alignment methods: the least squares method and the scale-based method. The results are shown in Table~\ref{tab:lse-scale}. The results indicate that the scale prediction can transform depths to a real-world scale. Subsequently, we visualize the real-world scale point clouds and camera extrinsics in Fig.~\ref{fig:lst-scale}. The results indicate that the real-scaled point clouds maintain the similar geometry consistency as the normalized ones.

\begin{table}[tb!]
\vspace{-3pt}
    \centering
    \caption{
        \textbf{Depth Estimation Method Comparison.} 
    }
    \vspace{-3mm}
    \linespread{1.3}
    \renewcommand\arraystretch{1}
    \begin{tabular}{ccc}
        \toprule
        {\multirow{3}{*}{\textbf{Method}}}&
        \multicolumn{2}{c}{\multirow{1}{*}{\textbf{frame=15}}}
        \\
        \cmidrule(r){2-3}
        & 
        {\multirow{1}{*}{Abs rel$\downarrow$}}&
        {\multirow{1}{*}{$\delta^3\uparrow$}}
        \\
        \midrule
        least squares method &0.3805 &0.8747\\
        scale-based method & 0.3666& 0.7412\\
        \bottomrule
    \end{tabular}
    \vspace{-1em}
    \label{tab:lse-scale}
\end{table}

\section{Conclusion}
In this work, we propose DriveVGGT, a feed-forward reconstruction model specializing in multi-camera geometry predictions. Compared to previous methods, DriveVGGT can effectively utilize relative camera poses to enhance the accuracy of geometric predictions, such as camera pose and depth estimation. Comprehensive evaluations on the nuScenes dataset demonstrate the outperforming performance compared to previous feed-forward methods, while maintaining lower computational consumption.

\section{Supplementary  Material}
\setcounter{page}{1}

\subsection{Further Details on Problem Formulation and Task Necessity}
Existing feed-forward reconstruction models fail to leverage three critical constraints inherent in Autonomous Driving (AD):

\begin{enumerate}
    \item Sparse camera overlap and high-resolution inputs make global attention mechanisms computationally inefficient.
    \item Fixed extrinsic calibrations (camera-to-vehicle) are not effectively utilized by previous methods.
    \item Most feed-forward frameworks fail to output poses, depths, and point clouds with absolute metric scale (meters or centimeters).
\end{enumerate}

To address these limitations, DriveVGGT is proposed for efficient, scale-aware 4D multi-camera reconstruction specifically tailored for AD scenes.

\subsection{Core Innovations of DriveVGGT}
\textbf{Bridging the "AD Gap":} Existing feed-forward methods struggle to incorporate fixed camera-to-vehicle geometry. DriveVGGT reformulates the pipeline by treating multi-camera extrinsics as fundamental geometric priors, significantly improving both accuracy and inference speed.

\textbf{Architectural Innovation:} To avoid the $O((M \times N)^2)$ complexity of standard global attention, our two-stage pipeline (TVA + MCA) decouples intra-camera temporal continuity from inter-camera geometric consistency. This enables scaling to long sequences without memory overflows (OOM), achieving a 2x speedup.

\textbf{Universal AD Framework:} DriveVGGT provides a systematic extension for feed-forward reconstruction in AD. Any geometry transformer (e.g., FastVGGT, VGGT) can be integrated into our pipeline to achieve efficient 4D reconstruction, effectively resolving the performance degradation inherent in multi-camera systems.

The above statement are concluded in Table~\ref{tab:innovation_comp} below:

\begin{table}[!hb]
    \centering
    \caption{\textbf{Comparison of DriveVGGT and Previous Models.}}
    \scalebox{0.85}{
    \renewcommand\arraystretch{1.2}
    \begin{tabular}{lcc}
        \toprule
        \textbf{Feature} & \textbf{Previous Models} & \textbf{DriveVGGT (Ours)} \\
        \midrule
        Data Modality & Sparse Images & Multi-camera Videos \\
        Geometric Priors & \xmark & \cmark~ (Fixed Extrinsics) \\
        Architecture & Global Attention & Decoupled TVA + MCA \\
        Framework Flexibility & \xmark & \cmark~ (Universal Plugin) \\
        \bottomrule
    \end{tabular}
    }
    \label{tab:innovation_comp}
\end{table}

\subsection{The Value of Ego-Vehicle Pose Estimation}
\textbf{Reliability Over External Sensors:} Unlike static camera-to-vehicle extrinsics, ego-motion data from GPS, IMU, or GNSS can be unstable or unavailable in challenging traffic environments. DriveVGGT provides a robust alternative by estimating these parameters directly through a feed-forward model, ensuring resilience when external sensor data fails.

\textbf{Relationship to Visual Odometry:} The value of ego-vehicle pose estimation shares similarities with visual odometry, a classic computer vision task in autonomous driving primarily addressing localization. Furthermore, DriveVGGT goes beyond pure ego-pose estimation by simultaneously predicting depth and achieving full scene reconstruction. By injecting static and precise multi-camera-to-vehicle poses, DriveVGGT effectively handles both multi-camera odometry and reconstruction tasks—capabilities not addressed by previous visual odometry or purely feed-forward reconstruction methods.

The above statement are concluded in Table~\ref{tab:pose_comparison} below:

\begin{table}[!ht]
    \centering
    \caption{\textbf{Comparison of different camera Poses.}}
    \scalebox{0.85}{
    \renewcommand\arraystretch{1.2}
    \begin{tabular}{lcc}
        \toprule
        \textbf{Attribute} & \textbf{Cam-to-Vehicle Poses} & \textbf{Vehicle-Ego Poses} \\
        \midrule
        Temporal Property & Static & Dynamic \\
        Robustness        & High (Fixed) & Low (sensor-dependent) \\
        Source            & Calibration & Sensors (GPS/IMU/GNSS) \\
        Accessibility     & Readily Available & Often Noisy/Intermittent \\
        Estimation Value  & Low (Used as Prior) & High (Core Output) \\
        \bottomrule
    \end{tabular}}
    \label{tab:pose_comparison}
\end{table}

\subsection{Pipeline Architectural Abstraction}
\begin{figure}[!ht]
    \centering
    \includegraphics[width=0.7\linewidth]{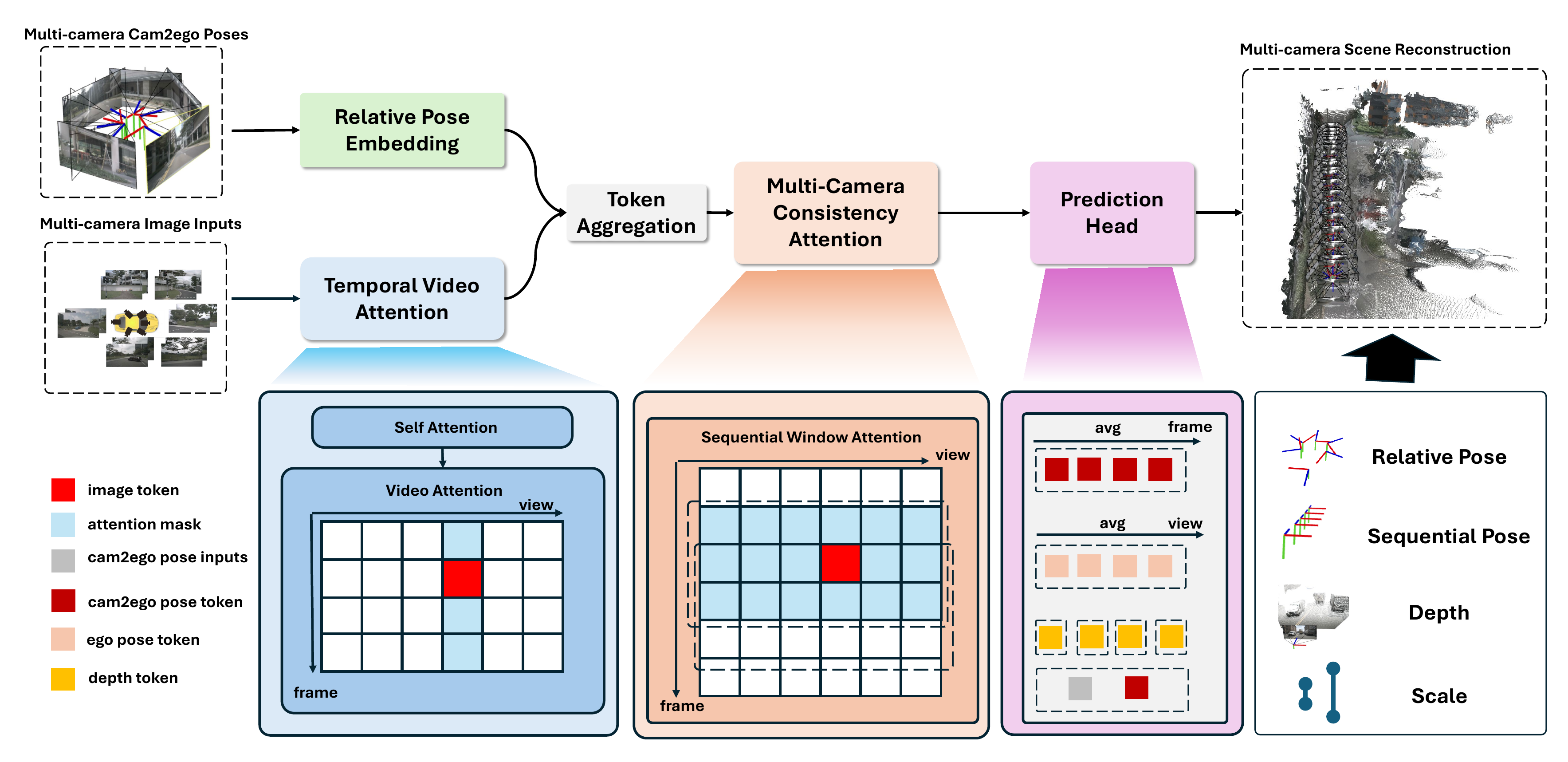}
    \caption{Detailed pipeline of DriveVGGT, highlighting the integration of the decoupled TVA and MCA modules.}
    \label{fig:rebuttal_pipeline}
\end{figure}

Figure~\ref{fig:rebuttal_pipeline} provides a concise  architectural abstraction of our pipeline, illustrating how the decoupled attention modules interface with the geometric priors.

\subsection{The Explanation of DriveVGGT (fastVGGT) Inference Time Degradation (compared with DriveVGGT (VGGT))}
In Table 5 of the manuscript, the inference time of  DriveVGGT (fastVGGT) is longer than that of DriveVGGT (VGGT), which may cause confusion. The explanation is that in DriveVGGT, we process images belonging to each camera separately (the function of the Temporal Video Attention module). Thus, the VGGT or fastVGGT treats only 1/6 of the images as a batch in the entire multi-camera inputs, and the image number of each batch is 15, 25, 35 (not 90, 150, 210). The inference time of VGGT and fastVGGT for different numbers of images is shown in the table~\ref{tab:reason}. The results indicate that VGGT is faster than fastVGGT when dealing with fewer images. That's why DriveVGGT (fastVGGT) is slower than DriveVGGT (VGGT).

\subsection{DriveVGGT Prediction Visualization}
\begin{figure}[!ht]
    \centering
    \begin{minipage}{0.52\textwidth}
        \centering
        \includegraphics[width=\linewidth]{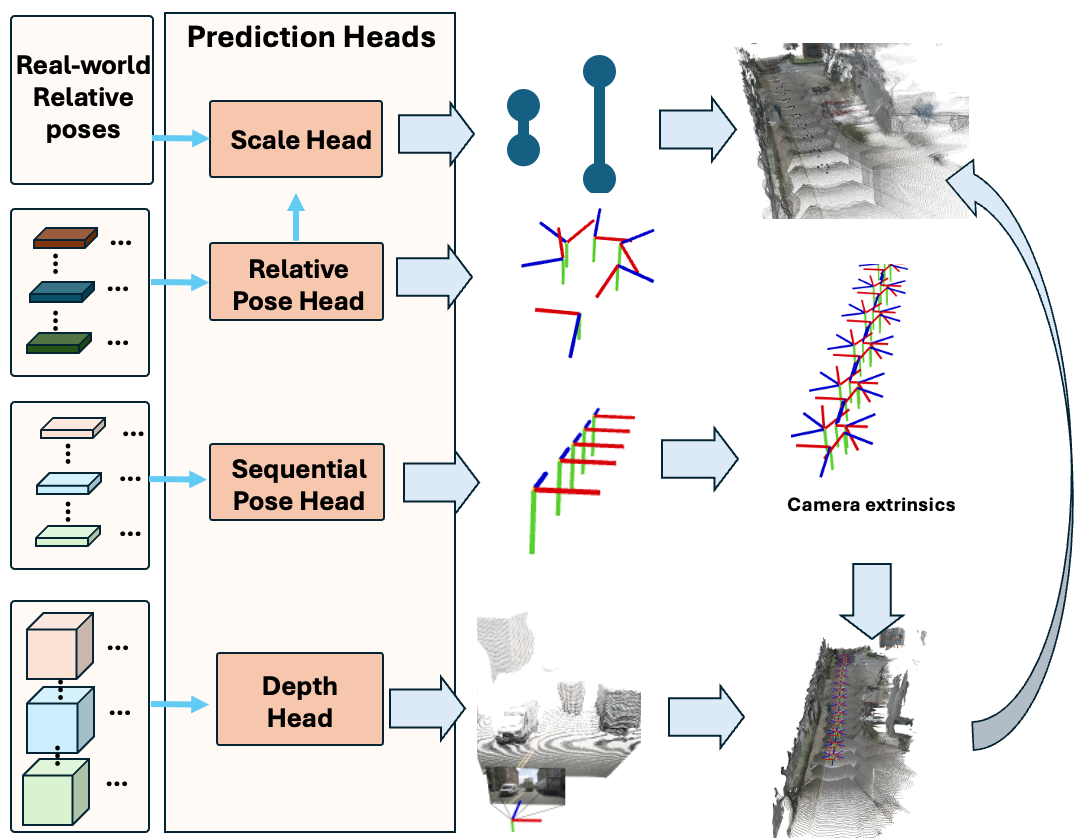}
        \caption{Prediction heads}
        \label{fig:predict}
    \end{minipage}
    \hfill 
    \begin{minipage}{0.40\textwidth}
        \centering
        \includegraphics[width=\linewidth]{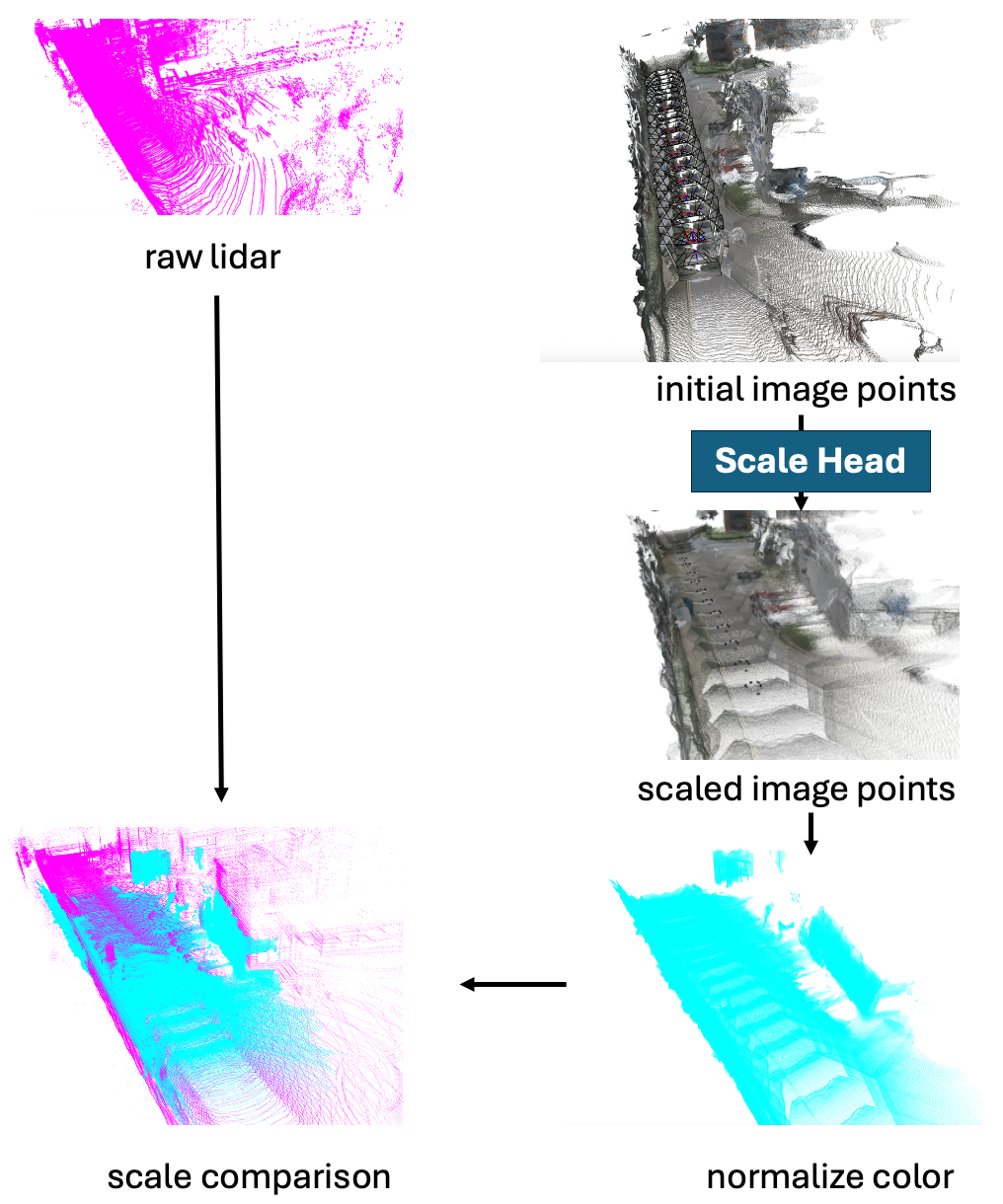}
        \caption{Comparison between raw lidar and scaled image points.}
        \label{fig:scale}
    \end{minipage}
\end{figure}

We visualize the DriveVGGT outputs in Fig~\ref{fig:predict} to better illustrate the relationship between prediction heads and tokens.

\begin{table}[!tb]
    \centering
    \caption{
        \textbf{Multi-camera inference time on nuScenes dataset.} 
    }
    \begin{tabular}{ccccc}
        \toprule
        {\multirow{3}{*}{\textbf{Method}}}&
        \multicolumn{4}{c}{\multirow{1}{*}{\textbf{Inference time $\downarrow$(ms)}}} 
        \\
        \cmidrule(r){2-5}
        & 
        {\multirow{1}{*}{images=6}}&
        {\multirow{1}{*}{images=30}}&
        {\multirow{1}{*}{images=54}}&
        {\multirow{1}{*}{images=150}}
        \\
        \midrule
        \textbf{VGGT}&\textbf{95}&\textbf{461}&\textbf{1027}&5241\\
        \textbf{fastVGGT}  &537&893&1277&\textbf{3341} \\
        
        \bottomrule
    \end{tabular}
    \vspace{-1em}
    \label{tab:reason}
\end{table}

\subsection{Scale Head Visualization}
To visualize the scaled global points from images, we visualize the scaled results and raw point clouds from the lidar sensor ~\ref{fig:scale}. To achieve better visualization results, we normalize the color of scaled image points. The scale comparison demonstrates that the model can achieve general real-world geometry accuracy with the scale head.

\subsection{More Visualization Results}

\begin{figure*}[h]
\centering
\includegraphics[width=0.78\linewidth]{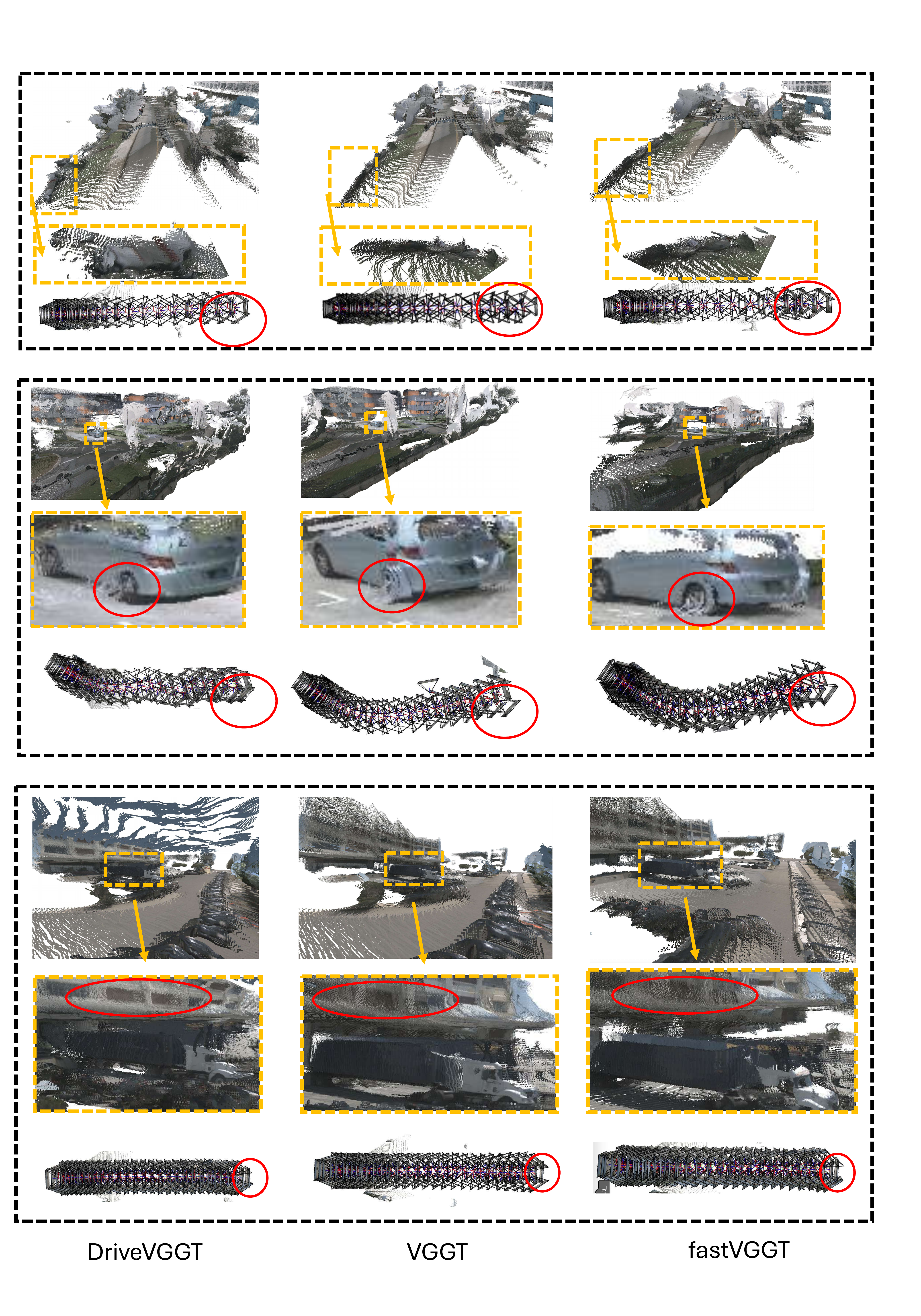}
\caption{Visualization Results 1.}
\label{figure}
\end{figure*}

\begin{figure*}[h]
\centering
\includegraphics[width=0.85\linewidth]{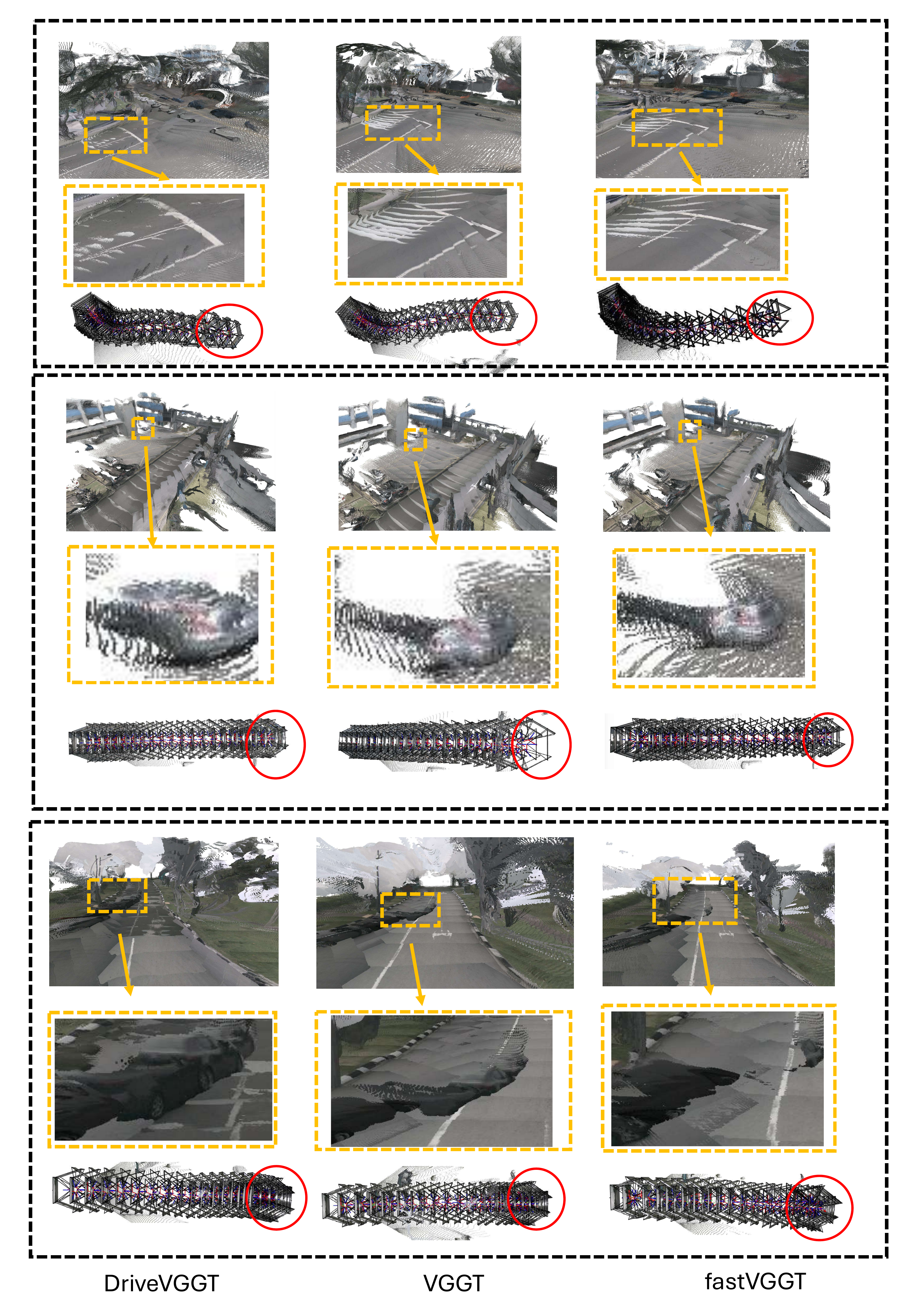}
\caption{Visualization Results 2.}
\label{figure}
\end{figure*}

\clearpage  


%
%
\bibliographystyle{splncs04}
\bibliography{main}
\end{document}